\documentclass[sigconf, nonacm]{acmart}
\AtBeginDocument{%
  }

\newif\ifshowcomments
\showcommentstrue

\usepackage{multirow}
\usepackage{enumitem}
\setlist[itemize]{topsep=3pt, itemsep=1pt, parsep=0pt, partopsep=0pt}
\usepackage{booktabs}
\usepackage[normalem]{ulem} 
\usepackage{booktabs,ulem}
\usepackage{xcolor}
\definecolor{darkgreen}{rgb}{0.0,0.5,0.0}
\definecolor{cvprblue}{rgb}{0.21,0.49,0.74}
\usepackage{booktabs}
\usepackage{tabularx}
\usepackage{makecell}
\usepackage{array}
\usepackage{pifont}

\usepackage{booktabs}
\usepackage{siunitx}
\usepackage[table]{xcolor}
\usepackage{subcaption}
\definecolor{frozencol}{gray}{0.965}
\definecolor{ftcol}{gray}{0.93}
\newcolumntype{F}{>{\columncolor{frozencol}}c}
\newcolumntype{T}{>{\columncolor{ftcol}}c}

\definecolor{groupgray}{gray}{0.90}
\definecolor{subgroupgray}{gray}{0.95}

\begin{document}

\title{How do Self-Supervised Remote Sensing Vision Models Transfer to Downstream Tasks?}

\author{Julia Romero}
\email{julia.romero@colorado.edu}
\affiliation{%
  \institution{University of Colorado Boulder}
  \city{Boulder}
  \state{CO}
  \country{USA}
}

\author{Qin Lv}
\email{qin.lv@colorado.edu}
\affiliation{%
  \institution{University of Colorado Boulder}
  \city{Boulder}
  \state{CO}
  \country{USA}
}

\author{Morteza Karimzadeh}
\email{karimzadeh@colorado.edu}
\affiliation{%
  \institution{University of Colorado Boulder}
  \city{Boulder}
  \state{CO}
  \country{USA}
}




\begin{abstract}
Self-supervised geospatial foundation models (GeoFMs) learn transferable representations from remote sensing data, but their downstream behavior is difficult to characterize. We study six representative GeoFMs spanning joint-embedding, reconstruction, and multimodal pretraining families, and evaluate transfer across classification, regression, and segmentation benchmarks under different label availability and downstream pipelines. We find that model rankings change across tasks and adaptation settings. Layerwise probing shows that, in most cases, task-relevant information is more accessible in intermediate transformer blocks compared to final-layer embeddings, and that GeoFMs exhibit distinct depthwise profiles. In segmentation case studies on PASTIS and Sen1Floods11, downstream adaptation settings such as decoder design and fine-tuning can be as impactful as the choice of GeoFM, and standard dense-prediction heads may be poorly aligned with how GeoFMs organize information over depth. Finally, CKA analysis on case studies shows that fine-tuning does not rewrite GeoFMs uniformly across depth, and the strongest changes are localized to the first linear layer of the MLP in ViT blocks. These results help explain why GeoFM rankings shift across benchmarks and motivate more representation-aware evaluation and adaptation strategies.



\end{abstract}   
\maketitle
 
\section{Introduction}
\label{sec:intro}

\begin{figure*}[t]
  \centering
  \includegraphics[width=\textwidth, trim={1pt 1pt 3pt 1pt},
    clip]{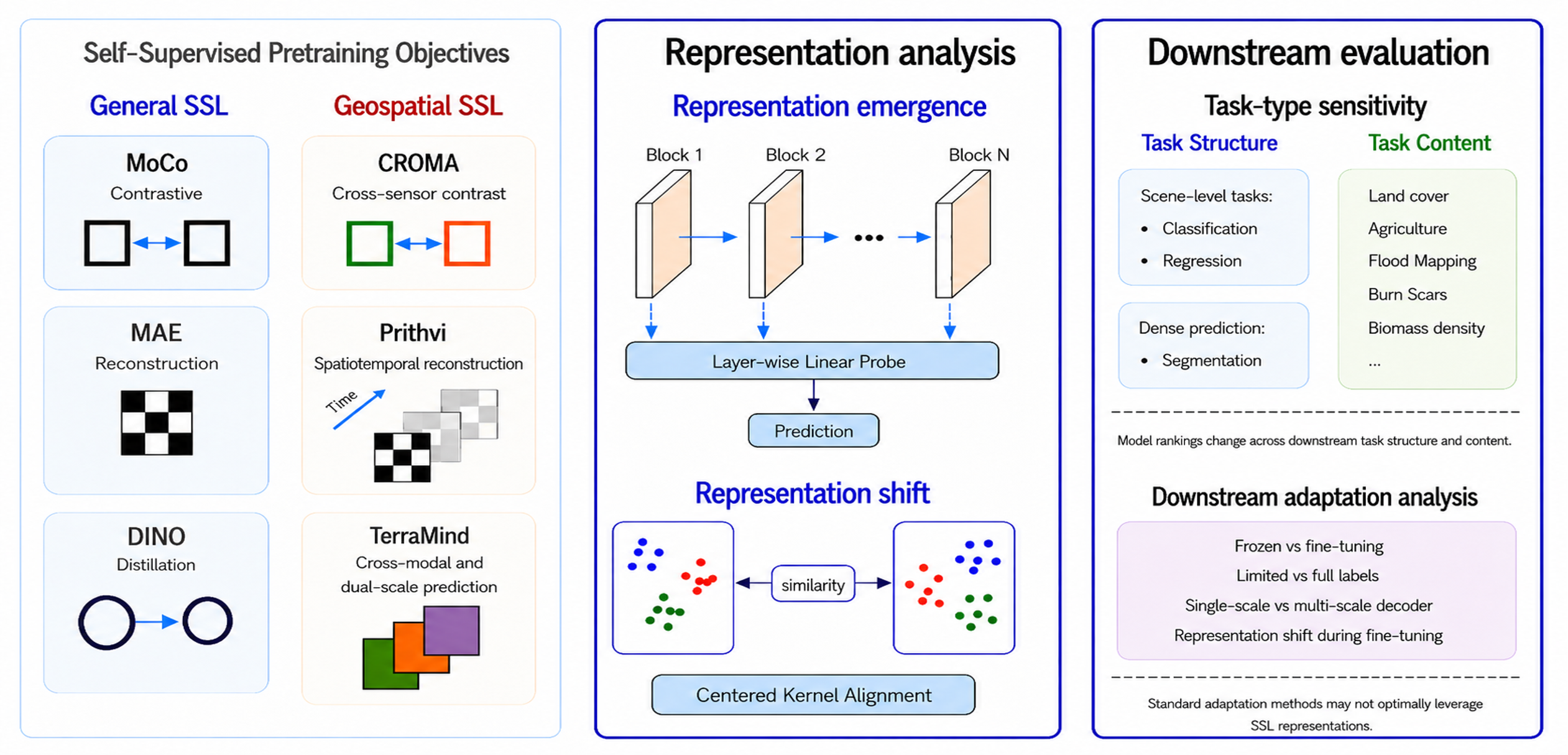}
  \caption{Overview of our study investigating  SSL objectives, characterizing representation learning, and evaluating downstream transfer conditions.}
  \label{fig:teaser}
\end{figure*}

Self-supervised learning (SSL) has enabled the development of remote sensing foundation models (GeoFMs) which leverage large-scale unlabeled satellite imagery. These models are trained to extract general-purpose representations of Earth Observation (EO) imagery and then fine-tuned or used as frozen feature extractors to adapt to diverse downstream tasks, such as crop monitoring, wildfire detection, and sea ice mapping \cite{rs15082092, JALAYER2025101538}. The strength of SSL models is their ability to perform on tasks with limited task-specific labeled training data, and this has been demonstrated in general computer vision through foundation models such as CLIP and DINO, which show impressive performance on diverse tasks \cite{radford2021learning, caron2021emerging}.

However, recent research shows that SSL methods adapted for remote sensing (RS) data have not produced the leap in performance seen in general computer vision, and for some tasks, GeoFMs lag behind supervised baselines \cite{rs17020179, gupta2024specialized, xiao2025foundationmodelsremotesensing, marsocci2025pangaeaglobalinclusivebenchmark}. Satellite data poses challenges that differ substantially from those in natural image understanding. Unlike natural images, remote sensing imagery often does not contain discrete objects with clear boundaries. Its structure is shaped by spatial resolution, spectral bands, geographic context, and repeated observations over time \cite{rolf2024mission}. Real-world geospatial applications also frequently require dense pixel-level prediction, where features must transfer to segmentation tasks rather than scene-level prediction alone \cite{yang2025survey}.

SSL models employ a variety of pretraining objectives and recent SSL objectives targeting the geospatial domain aim to leverage structure inherent in EO data, adapting these methods to spectral, temporal, and geospatial dimensions. Recent benchmarks such as PANGAEA-Bench \cite{marsocci2025pangaeaglobalinclusivebenchmark} have begun evaluating these models on dense, pixel-level prediction tasks with varying multispectral channels, image resolutions, and temporality, enabling a more comprehensive picture of transfer to real-world tasks. Yet across these benchmarks, no single design consistently dominates across downstream tasks \cite{marsocci2025pangaeaglobalinclusivebenchmark, yang2025survey, xiao2025foundationmodelsremotesensing}. Differences in pretraining datasets, architecture scale, and benchmarking frameworks partially account for this inconsistency, making it difficult to compare pretraining frameworks \cite{huang2025survey}.

 Different SSL objectives may preserve different kinds of information about the input. Representations are only effective during inference if they integrate robust priors that generalize \cite{odonnat2026layer}. While recent publications have benchmarked GeoFM performance, there is limited work that examines how different SSL pretraining objectives transfer differently across downstream tasks. For example, two models with similar accuracy on a benchmark may differ substantially in the spatial, spectral, or semantic information they preserve, which affects how well each transfers to different tasks. Furthermore, downstream performance depends on whether the evaluation and adaptation strategy can access the relevant information. In addition to performance, it is important to study what information each pretraining objective preserves, how the information is extracted over network depth, and how downstream methods utilize it. Standard transfer heads may not be well matched to how task-relevant information is organized over depth in GeoFMs.   

In this work, we characterize how SSL models with common vision transformer (ViT) backbones extract representations from remote sensing images. We investigate how pretrained SSL encoders transfer to downstream tasks under different adaptation conditions, including frozen-encoder probing versus fine-tuning, varying label availability, and decoder designs. Specifically, we examine a set of general SSL models trained on remote sensing datasets alongside geospatial SSL models which explicitly leverage structure of geospatial data, spanning contrastive, distillation, reconstruction, and multi-modal training objectives. Our contributions are as follows:
\begin{itemize}
    \item \textbf{Downstream transfer evaluation across task types.} We evaluate six representative SSL GeoFMs across diverse downstream tasks including image-level classification and regression and pixel-level segmentation, with varying label availability. Our results show that model rankings are highly sensitive to each condition examined.

    \item \textbf{Depthwise Analysis of SSL representations.} Using layerwise probes, we demonstrate that self-supervised objectives organize low-level and semantic information differently across network depth, and intermediate representations can be more informative than final-layer embeddings.

    \item \textbf{Analysis of downstream adaptation and fine-tuning.} Downstream adaptation choices, including frozen vs. fine-tuned encoders and decoder design, substantially influence transfer performance, and lighter decoders often match or outperform heavier multi-scale commonly-used designs. We show that fine-tuning induces model-specific depthwise layer updates, guiding future work towards GeoFM adaptation.

\end{itemize}

Our findings show that transfer behavior of GeoFMs depends on the pretraining objective, downstream task, and adaptation strategy. Through examining each of these factors, our contributions can guide development of downstream transfer schemes and how future SSL objectives and adapters may be designed for more accurate and efficient Earth observation.

\section{Related Work}
\label{sec:related}

\subsection{Natural versus Remote Sensing Imagery} \label{differences}
Computer vision architectures developed for natural images are often applied to RS imagery, but the assumptions underlying these models may not transfer cleanly to RS data, which differ in structure and content. The natural image domain and adaptation modules focus on object- and scene-level recognition, and features at different depths of the network capture textures, parts, objects, and scene understanding \cite{xiao2018unifiedperceptualparsingscene}. RS tasks may rely more on spectral, spatial, temporal, and geographic context (i.e., depend on where on the Earth the feature is located). Additionally, there are RS specific properties impacted by environmental factors, such as brightness, temperature, and season \cite{rs17020179}. A single class may range from one pixel to covering the full image, and remote sensing data also has a temporal dimension, which may be necessary to capture the latent information of the Earth patch \cite{rolf2024mission}. Brown et al. show that an architecture almost 100× smaller than common natural image segmentation models achieves state-of-the-art performance on land cover mapping \cite{brown2022dynamic}, suggesting that natural image techniques may not be suitable for some RS applications. Overall, it is necessary to understand how deep learning models extract representations from RS imagery and whether downstream adapters leverage representations appropriately \cite{rolf2024mission, zhu2026foundations}.

\subsection{Self-Supervised Learning for Natural Computer Vision}
General vision models and remote sensing variants are commonly pretrained with two families of self-supervised objectives: (i) joint embedding methods, including contrastive learning and self-distillation, which align representations across different views of the same location, and (ii) reconstruction methods with masking, which learn by reconstructing masked elements of the input image. Canonical SSL methods in the natural vision domain include SimCLR, CLIP, and MoCo (contrastive), DINO (distillation), and MAE (reconstruction)  \cite{chen2020simple, radford2021learning, he2020momentum, caron2021emerging, he2022masked}. These models achieve strong performance on zero-shot prediction on natural images. These foundational self-supervised methods inspire several geospatial variants which adapt self-supervised objectives to EO pretraining datasets \cite{lane2026genealogyfoundationmodelsremote, wang2023ssl4eo, liu2024remoteclip}.

\subsection{Geospatial SSL Pretraining Objectives}
More recent methods exploit structure specific to Earth observation data, extending these objectives to temporal sequences, multispectral channels, cross-modality interactions, and multi-scale spatial structure \cite{manas2021seasonal, fuller2023croma, cong2022satmae, reed2023scale, tseng2502galileo, jakubik2025terramind, jakubik2023foundationmodelsgeneralistgeospatial, guo2024skysensemultimodalremotesensing}. Prior work suggests that globally contrastive methods learn features that align with scene-level semantic tasks but can be less effective for tasks requiring fine-grained spatial detail, such as object-localization within an image \cite{wang2021dense, Pang_2024_WACV}. This shortcoming is relevant to geospatial domain where dense prediction tasks are high priority, such as pixel-wise semantic segmentation (e.g., sea ice or land cover mapping) to avoid mosaicing patch classification outputs and the potential resulting artifacts. To address this, recent geospatial models incorporate local or multi-scale contrastive signals within a region, typically at higher computational costs. For example, SkySense \cite{guo2024skysensemultimodalremotesensing} applies cross-modal contrastive learning across pixel-, region-, and image-level representations, and Galileo \cite{tseng2502galileo} couples multi-scale contrast and masking (without reconstruction) using contrastive loss applied at the pixel-level and image-level. 

In comparison, reconstruction methods prioritize within-image local structure, but may be less effective on tasks that depend strongly on scene semantics. Reconstruction approaches in geospatial vision models adapt masking to the unique structure of satellite imagery. For example, SatMAE \cite{cong2022satmae} applies masking across temporal samples of the same geolocation, Scale-MAE \cite{reed2023scale} introduces scale-aware masking/reconstruction to learn relationships invariant to changing spatial scales and resolutions. Similarly, Prithvi extends the general MAE architecture, applying masked reconstruction to temporal inputs to learn spatiotemporal representations \cite{jakubik2023foundationmodelsgeneralistgeospatial}.

Other methods apply these objectives to multimodal satellite data. For example, CROMA couples cross-modal contrastive learning with reconstruction and shows that the contrastive signal is more vital than the reconstruction signal \cite{fuller2023croma}. TerraMind is another recent state-of-the-art model which employs cross-modal token prediction to predict masked information across a large set of EO modalities in addition to a pixel-token dual scale objective \cite{jakubik2025terramind}. 

Despite proliferation of GeoFM research, currently they are not suitable for ubiquitous practical use without substantial fine-tuning \cite{rs17020179}. There is a wide range of dataset and model complexity scales, and larger frameworks require significant compute for training and deployment which is also a challenge for practical use \cite{rs17020179, eotowards}.


\subsection{Evaluation and Interpretation of Remote Sensing SSL Representations}
GeoFM evaluation is commonly based on downstream benchmark performance or final-layer embeddings and may use standardized adaptation protocols across models. Prior work shows that GeoFMs do not consistently outperform supervised baselines on dense prediction (segmentation), even in a 10\% limited-label setting \cite{marsocci2025pangaeaglobalinclusivebenchmark, simumba2026geobench2performancecapabilityrethinking}. Models with similar downstream performance can encode task-relevant information at different depths \cite{shekhar2023objectivesmatterunderstandingimpact}. Prior studies of representation learning in natural vision have shown that SSL can produce structured stages of learning across depth, including emergent qualities aligned with downstream tasks \cite{caron2021emerging}. Analyzing where geospatial information becomes accessible in GeoFM encoders is important for both model selection and downstream adaptation.


Recently \cite{martiescofet2025finetunesmarterharderparameterefficient} evaluated parameter-efficient fine-tuning methods for GeoFMs and demonstrate that low-rank adaptation (LoRA) can match or exceed full fine-tuning, and decoder  substantially affects segmentation performance. \cite{gilch2026embedmattersevaluationeo} has examined GeoFM transfer behavior regarding representation depth, embedding aggregation, and backbone architecture in frozen-encoder settings, showing that combining embeddings from different objectives can improve robustness. Corley et al. \cite{corley2026pixelspatchespoolingstrategies} demonstrate that token pooling strategy impacts generalization across geographic regions. Further work is needed to understand how representations are extracted, utilized, and adapted to downstream tasks \cite{zhu2026foundations}. In this work, we build on this line of transfer analysis by applying layerwise linear probes and centered kernel alignment (CKA) to characterize how SSL objectives shape geospatial representations and how those representations change during downstream fine-tuning.

\noindent\textbf{Linear Probe} 
In natural imagery computer vision, linear classifier probes have been used to identify where task-relevant information becomes linearly accessible and can be used to measure how representations are transformed through network depth \cite{alain2018understandingintermediatelayersusing, sanghavi2026edges}. Insights can suggest more effective strategies for downstream transfer; for example, \cite{odonnat2026layer} shows that intermediate layer representations may be more robust to distribution shift than final layer \cite{odonnat2026layer}.  


\noindent\textbf{Representation Space Similarity}
Centered Kernel Alignment (CKA) is a measure of similarity between representation spaces and has been used to compare depth-wise features across and within neural networks \cite{kornblith2019similarity, shekhar2023objectivesmatterunderstandingimpact, raghu2021vision}.




\section{Methodology}
\label{sec:framework}



\begin{table*}[h]
\centering
\footnotesize
\setlength{\tabcolsep}{5pt}
\renewcommand{\arraystretch}{1.18}
\begin{tabularx}{\textwidth}{
>{\raggedright\arraybackslash}p{3cm}
>{\centering\arraybackslash}p{2cm}
>{\centering\arraybackslash}p{2cm}
>{\raggedright\arraybackslash}X
>{\raggedright\arraybackslash}p{4.7cm}
}
\toprule
\textbf{Model} & \textbf{\makecell{Geospatial Prior}} & \textbf{Modalities} & \textbf{SSL objective} & \textbf{Architecture} \\
\midrule

\textbf{MoCo (SSL4EO)} \cite{wang2023ssl4eo} &
None &
Optical &
Contrastive &
Momentum encoder + queue \\

\textbf{MAE (SSL4EO)}  \cite{wang2023ssl4eo} &
None &
Optical &
Masked reconstruction &
Spatial encoder--decoder \\

\textbf{DINO v1 (SSL4EO)}  \cite{wang2023ssl4eo} &
None &
Optical &
Distillation &
Student--teacher \\

\midrule
\rowcolor{black!4}
\textbf{Prithvi v1} \cite{jakubik2023foundationmodelsgeneralistgeospatial} &
Temporal structure&
Optical &
Masked reconstruction &
Spatiotemporal encoder--decoder \\
\rowcolor{black!4}
\textbf{CROMA} \cite{fuller2023croma} &
Cross-sensor &
Optical--SAR &
Hybrid (contrastive + masking) &
Dual encoders + cross-modal attention + shared decoder \\
\rowcolor{black!4}
\textbf{TerraMind} \cite{jakubik2025terramind} &
Cross-modal  &
Multimodal &
Masked token prediction and pixel-token dual scale learning &
Unified multimodal encoder--decoder with modality-specific tokenizers \\
\bottomrule
\end{tabularx}
\vspace{2pt}
\caption{Model suite used in our study grouped by general-purpose SSL versus geospatial-specific SSL techniques. Optical indicates multispectral inputs from Sentinel-2 or HLS (Prithvi). }
\label{tab:model_suite_objectives}
\end{table*}

\subsubsection{Model Selection}
We select a set of available pretrained models representing both general-purpose vision SSL backbones and geospatial-specific pre-trained models, spanning contrastive, distillation, reconstruction, and multimodal pretraining objectives. For general SSL methods we evaluate momentum contrast (MoCo), distillation (DINO v1), and masked autoencoder (MAE), all pretrained on the SSL4EO dataset \cite{wang2023ssl4eo}. For the geospatial-prior subset, we select recent state-of-the-art models: Prithvi v1 (100M) \cite{jakubik2023foundationmodelsgeneralistgeospatial}, CROMA \cite{fuller2023croma}, and TerraMind \cite{jakubik2025terramind}, which represent temporal reconstruction, cross-sensor alignment, and multimodal pretraining objectives, respectively.

To compare the evolution of representations over depth, we use backbones with 12 ViT blocks for all models. This corresponds to ViT-S/16 for the general SSL models of MoCo, DINO, and MAE, and ViT-B/16 for the geospatial-specific models of Prithvi and TerraMind. CROMA uses ViT-B/8 but downsamples images to roughly halved resolution of 120x120 compared to 224x224 used for the other models. Prithvi v1 is used because Prithvi v2 does not release weights corresponding to a 12 block ViT. Although all backbones have the same number of transformer blocks, the ViT-S models use half as many attention heads and substantially fewer parameters than the ViT-B models, resulting in backbone sizes of approximately 23M and 88M parameters, respectively.

\textbf{Table ~\ref{tab:model_suite_objectives}} shows details on the data modalities, objectives, and architectures of each model in the suite. All models except for Prithvi and TerraMind were pretrained on SSL4EO. Prithvi is trained on the NASA optical HLS dataset which has 4.2$\times$ samples as SSL4EO \cite{jakubik2023foundationmodelsgeneralistgeospatial, wang2023ssl4eo}, and TerraMind is trained on TerraMesh, which has 9$\times$ samples as SSL4EO and contains diverse modalities. \textbf{Table~\ref{tab:model_suite_scale}} summarizes the scale of each model's pretraining framework. MoCo pretraining framework is the smallest with 42M parameters, and TerraMind possesses the most at 2B, almost 50$\times$ more than MoCo. Given the range of training framework scale, it is worth investigating if these design and training data choices which require substantially more compute translate to consistent improvements in performance. Accordingly, we also evaluate a ViT-B/16 baseline trained with supervision on ImageNet and the first layer is adapted to accept 13 input channels. The RGB weights were used for RGB channels and repeated to initialize the other 10 channels.

\begin{table}[h]
\centering
\footnotesize
\setlength{\tabcolsep}{3pt}
\renewcommand{\arraystretch}{1.1}
\begin{tabular}{l c c c l}
\toprule
\textbf{Model} & \textbf{Backbone} & \textbf{Pretrain Params} & \textbf{$\uparrow$ Params} & \textbf{Pretrain Data} \\
\midrule
\textbf{MoCo}       & ViT-S/16         & 42M      & \textbf{1.0$\times$}  & 1M, SSL4EO \\
\textbf{MAE}        & ViT-S/16         & 50M      & 1.2$\times$  & 1M, SSL4EO \\
\textbf{DINO v1}       & ViT-S/16         & 90M      & 2.1$\times$  & 1M, SSL4EO \\
\textbf{Prithvi v1} & ViT-B/16         & 113M     & 2.7$\times$  & 4.2M, NASA HLS \\
\textbf{CROMA}      & ViT-B/8         & 198M     & 4.7$\times$  & 1M, SSL4EO \\
\textbf{TerraMind}  & ViT-B/16 & 2,084M   & 49.6$\times$ & 9M, TerraMesh \\
\hline
\bottomrule
\end{tabular}

\caption{Scale of pretraining frameworks in the model suite. All inference encoders are 12-block ViTs. Train Params and Pretrain Data indicate the scale of each pretraining framework. During inference, only Backbone encoders are used, so parameters are comparable within ViT-S and ViT-B groups.}
\label{tab:model_suite_scale}
\end{table}

\begin{table*}[h]
\centering
\small
\setlength{\tabcolsep}{5pt}
\renewcommand{\arraystretch}{1.12}
\begin{tabular}{p{2.5cm} p{1.7cm} p{7cm} cc}
\toprule
\textbf{Decoder/ Module} & \textbf{Hooks} & \textbf{Method} & \textbf{ViT-S, ch=512} & \textbf{ViT-B, ch=512} \\
\midrule
UPerNet \cite{xiao2018unifiedperceptualparsingscene}
& Multi-level
& Multi-scale FPN using multiple block features and apply hierarchical convolutions with Pyramid Pooling Module
& 30.90M & 39.35M \\

Lightweight Multi-Scale Fusion
& Multi-level
& Multi-scale FPN using multiple encoder blocks, concatenate, 3x3 single convolution 
& 12.01M & 18.11M \\

Single-Level Upsampler
& Single-level
& 4 transposed conv upsampling blocks
& 1.53M & 6.20M \\
\bottomrule
\end{tabular}
\caption{Parameter counts and summary of the decoder architectures used in our segmentation experiments.}
\label{tab:decoder_summary}
\end{table*}

\subsubsection{Benchmarking Datasets}
To test whether transfer behavior depends on task structure, we evaluate the optical encoder using several benchmark frameworks spanning diverse tasks in classification, regression, and segmentation. Unless otherwise noted, downstream evaluations use frozen optical encoders. We use only optical (Sentinel-2) imagery in this work.

We use EuroSAT \cite{helber2019eurosatnoveldatasetdeep}, a standard benchmark for land-cover classification, using $k$NN with cosine similarity to evaluate the quality of the embedding space. We use NeuCo-Bench \cite{vinge2025neuco} for image-level regression, which includes both low-level tasks (e.g., biomass, heat-island, and cloud-related quantities) and higher-level semantic tasks (e.g., crop, agriculture, and forest coverage). For NeuCo-Bench, we use the default settings and released code\footnote{https://github.com/cmalbrec/NeuCo-Bench}, except that layerwise probes use 20-fold cross-validation instead of the default 40-fold due to compute limitations. Following \cite{vinge2025neuco}, we average per-season embeddings after feature extraction. For EuroSAT and NeuCo-Bench evaluations, images are normalized and preprocessed with the same method used for pretraining each model, since these use light kNN and linear regression heads.

We use PANGAEA-Bench \cite{marsocci2025pangaeaglobalinclusivebenchmark} code repository\footnote{https://github.com/VMarsocci/pangaea-bench/} for semantic segmentation tasks, using default settings for each task, including limited-label stratified sampling, normalization, band padding, and training hyperparameters. Specifically we employ datasets for disaster mapping (HLSBurnScars \cite{HLS_Foundation_2023} and Sen1Floods11 \cite{bonafilia2020sen1floods11}), marine oil spill mapping (MADOS) \cite{KIKAKI202439}, and agricultural mapping (AI4SmallFarms) \cite{1f1a4724360f45b08184b244e644b728}. These datasets contain variable image resolutions, number of classes, and band sets. For these tasks, we use only the 10\% label setting due to compute constraints.

For the case studies we use PASTIS \cite{garnot2021panoptic} with 10\% labels, which is a multi-temporal crop type segmentation dataset of Sentinel-2 image time series over France, with 18 crop classes and 6 timesteps per patch \cite{garnot2021panoptic}.  The standard configuration uses the L-TAE temporal aggregation module which passes an aggregated representation to the segmentation decoder. We also use both 10\% and 100\% labels with Sen1Floods11 which consists of binary water/no-water labels for 11 flood events globally \cite{bonafilia2020sen1floods11}. We select these tasks as diverse case studies because they represent a complex multi-class and temporal problem focused in a single-geographic region and a simpler binary-class spanning geographic regions.

\subsection{Layerwise Probe of Representations} \label{low-vs-high}
By evaluating the intermediate representations on specific downstream tasks, we can estimate which depths of the network contain information more relevant to that task. We probe intermediate blocks of ViT encoders by attaching a single trainable linear regression layer to selected blocks (frozen). We select tasks from the NeuCo-Bench dataset which are grouped into \textbf{Low-Level} and \textbf{High-Level/Semantic} tasks (following \cite{vinge2025neuco}). 

Low-level tasks are closely linked to sensor-level signals, while semantic tasks require more abstract scene-level understanding. For instance, estimating biomass density is considered a low-level task because it is commonly estimated with standard indices such as the NDVI (Normalized Difference Vegetation Index), a linear combination of spectral bands. High-level scene understanding is not necessary, although it may help predictions through other mechanisms such as handling geospatial artifacts mentioned previously. In contrast, semantic tasks such as crop coverage require distinguishing different land-cover types within the image.

\subsection{Case Studies: Downstream Transfer and Adaptation} \label{fine-tune}
\subsubsection{Downstream Transfer Configurations}
To study how pretrained GeoFM representations adapt during downstream learning, we evaluate case studies under different adaptation conditions including temporal aggregation, decoder design, and fine-tuning. We compare a \textit{frozen encoder + trainable decoder} setting against an \textit{end-to-end fine-tuned} setting.

We compare three decoder designs for segmentation with single-scale and multi-scale structure, and ranging from 1.5M parameters to 40M parameters. These decoders are detailed in Table~\ref{tab:decoder_summary}. We use UPerNet, a multi-scale decoder, which is common in geospatial dense prediction applications including in PANGAEA-Bench \cite{cha2023billion, marsocci2025pangaeaglobalinclusivebenchmark}. UPerNet was originally designed for scene understanding on natural images, where multi-scale fusion is suitable for hierarchical recognition of textures, parts, and objects at different spatial scales \cite{xiao2018unifiedperceptualparsingscene}. We additionally evaluate a lightweight multi-scale decoder that projects each hooked encoder layer to a common channel width, resizes all feature maps to the highest spatial resolution, concatenates them, and applies a single 3×3 convolution to fuse scales These multi-scale designs fuse multiple intermediate layer features. Finally, we apply a layerwise segmentation decoder which applies sequential transposed convolutions to upsample features from one layer. This is the lightest decoder, using 1.5-6.2M parameters (ViT-S to ViT-B).

\subsubsection{Measuring Representation Shift}
To quantify representational change induced by fine-tuning, we compute centered kernel alignment (CKA) using a PyTorch implementation of CKA-similarity\footnote{https://github.com/jayroxis/CKA-similarity} based on \cite{kornblith2019similarity}. For each pretrained/fine-tuned model pair, we extract activations from six hook points per ViT block, capturing activations spanning attention, feed-forward, layer-normalization, and residual connections. These hooks were selected as representative layers distributed throughout the ViT block, following \cite{raghu2021vision}.  

Patch-token representations are mean-pooled before CKA computation. We then compute same-layer CKA between matched hook activations from the pretrained and fine-tuned models, using the fine-tuning preprocessing and normalization pipelines. We used 2{,}500 images sampled deterministically across runs from either PASTIS or Sen1Floods11 to generate CKA results for fine-tuned encoders on each, respectively.




\section{Results}
\label{sec:experiments}

\begin{table*}[h]
\centering
\footnotesize
\setlength{\tabcolsep}{3.5pt}
\renewcommand{\arraystretch}{1.12}
\resizebox{\textwidth}{!}{%
\begin{tabular}{lccccccccccc}
\toprule
& \multicolumn{4}{c}{\textbf{PANGAEA 10\% Segmentation (mIoU)}} 
& \multicolumn{4}{c}{\textbf{NeuCo Regression ($R^2$)}} 
& \multicolumn{3}{c}{\textbf{EuroSAT kNN Accuracy}} \\
\cmidrule(lr){2-5} \cmidrule(lr){6-9} \cmidrule(lr){10-12}
\textbf{EO Model}
& \textbf{Farms}
& \textbf{BurnScars}
& \textbf{MarineDebris}
& \textbf{Floods}
& \textbf{10\% Low}
& \textbf{10\% Sem}
& \textbf{100\% Low}
& \textbf{100\% Sem}
& \textbf{1-shot (L2)}
& \textbf{10\% L2}
& \textbf{10\% Cos} \\
\midrule
MoCo
& \textbf{37.08} & 77.84 & 27.40 & \underline{84.34}
& -0.03 & 0.57 & 0.31 & 0.79
& \underline{73.54} & \textbf{93.98} & 74.11 \\
MAE
& 23.60 & 75.38 & 33.87 & 78.76
& 0.07 & 0.60 & \underline{0.40} & 0.73
& 65.32 & 88.35 & 75.56 \\
DINO
& 34.87 & 74.55 & 32.28 & 80.71
& -1.65 & 0.65 & 0.20 & 0.85
& \textbf{78.55} & 92.74 & 76.22 \\
Prithvi
& 36.25 & \textbf{78.88} & 29.92 & 83.62
& \textbf{0.25} & 0.55 & \underline{0.40} & 0.74
& 52.77 & 78.35 & 75.02 \\
CROMA
& \underline{36.42} & \underline{78.28} & \textbf{40.21} & 84.03
& -0.85 & \underline{0.71} & 0.29 & \underline{0.86}
& 71.00 & 90.67 & \underline{82.44} \\
TerraMind
& 12.54 & 77.05 & \underline{35.90} & \textbf{85.75}
& \underline{0.15} & \textbf{0.83} & \textbf{0.54} & \textbf{0.90}
& 70.34 & \underline{93.56} & \textbf{83.41} \\
\hline
ImageNet ViT-B/16 (13 ch)
& 32.27 & 71.70 & 18.01 & 79.26
& -0.25 & 0.51 & 0.19 & 0.67
& 45.71 & 71.57 & 70.72 \\
\bottomrule
\end{tabular}%
}
\caption{Frozen encoder downstream evaluations across segmentation, regression, and classification tasks. Model rankings are highly inconsistent across tasks and label abundance.  NeuCo columns report averaged $R^2$ over low-level tasks (biomass mean/std, clouds, heat-island mean/std) and semantic tasks (crops, land-cover agriculture, land-cover forest). EuroSAT reports 1-NN classification accuracy on 100\% split, while 10\% columns use 20-NN using Euclidean (L2) or cosine distance. }
\label{tab:representative_downstream_tasks}
\end{table*}

\subsection{Sensitivity of GeoFM Rankings to Task Type}
First, we present results across several downstream tasks to highlight the sensitivity of downstream evaluations of GeoFMs. Table~\ref{tab:representative_downstream_tasks} compares frozen-backbone performance across image-level tasks (classification and regression) and dense prediction tasks (semantic segmentation).  The results show that model rankings shuffle across the panel of diverse tasks, which include crop prediction, disaster mapping, land cover classification, and biomass detection. On regression tasks and EuroSAT, model rankings change across label availability, and on EuroSAT they also change across different distance metrics for kNN. The ImageNet baseline is generally not strongly competitive with the GeoFM model suite in these frozen encoder evaluations. Refer to Section \ref{imagenet} for discussion of ImageNet baseline performance.

\subsubsection{Task-related factors affecting transfer.}
Our findings align with recent work showing the sensitivity of GeoFM rankings to downstream applications \cite{marsocci2025pangaeaglobalinclusivebenchmark, simumba2026geobench2performancecapabilityrethinking}, and we additionally show inconsistent performance rankings for scene-level classification and regression alongside dense prediction performance across multiple models. \cite{marsocci2025pangaeaglobalinclusivebenchmark, simumba2026geobench2performancecapabilityrethinking} discuss factors related to downstream dataset characteristics including varying image resolutions, sensor channels, temporal configurations, and geographic coverage. These dataset-related factors may disproportionately benefit models pretrained with similar data or architecture. Due to these factors, it is difficult to isolate the impact of the SSL objective choice on downstream performance; however, in our study there is still high sensitivity among the three general-SSL models that were pretrained on the same dataset (SSL4EO) and ViT-S backbone. Next, we discuss links between SSL objective families and downstream task performance.

\subsubsection{Impact of pretraining objective on downstream transfer.}
Based on controlled prior work \cite{shekhar2023objectivesmatterunderstandingimpact, gilch2026embedmattersevaluationeo}, we hypothesize that different SSL pretraining objectives learn representation spaces that are better suited for different types of geospatial tasks. MoCo and DINO (joint embedding) are quite strong on high-level EuroSAT but are relatively weaker on NeuCo Semantic and Low-Level, suggesting that their final embeddings may not retain fine information relevant to regression tasks. Additionally, we observe that MAE and Prithvi, which employ reconstruction-based objectives, rank near the top on low-level tasks but score low on semantic tasks (NeuCo Semantic and EuroSAT). TerraMind and CROMA, models trained with cross-modal multi-objectives, outperform all other models on Neuco Semantic regression. Finally, the segmentation results show models are highly sensitive to segmentation task, with TerraMind falling short on AI4SmallFarms, despite its strong performance across the other tasks. Our results highlight that the examined models with joint embedding and reconstruction objectives prioritize semantic or low-level information, respectively, and methods like CROMA that combine both objectives are promising for controlling this balance and tuning the final embedding space. 


While EuroSAT and NeuCo tasks are image-level and use simple prediction heads (linear probe and kNN), dense prediction tasks commonly use a heavier decoder head that combines multiple intermediate layers. These methods access information from the encoders differently. In the following sections, we examine where task-relevant information becomes accessible across depth and interaction with downstream adaptation modules more closely.

\begin{figure*}[h]
    \centering
        \includegraphics[width=0.9\linewidth]{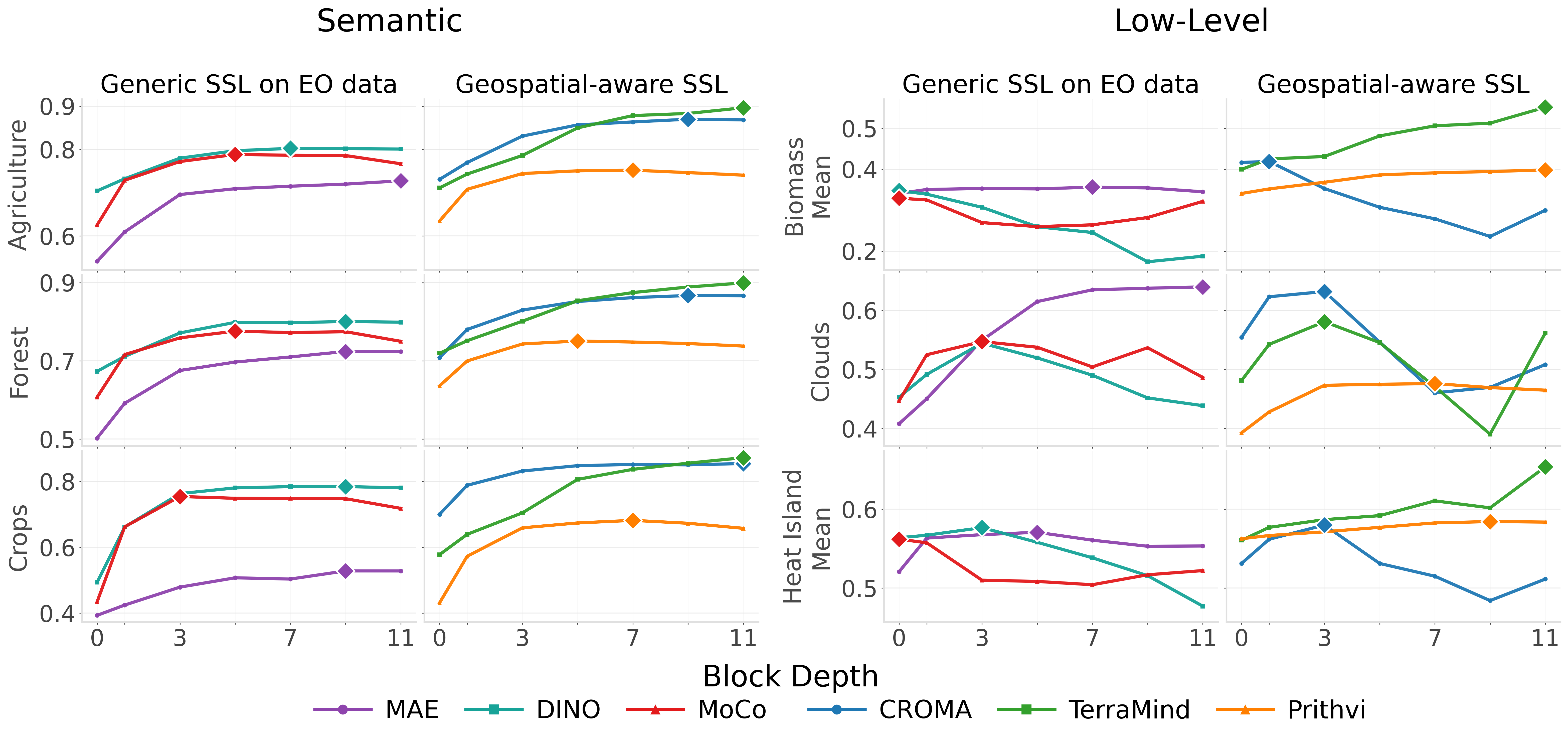}

    \caption{Linear-probe $R^2$ using intermediate ViT block representations to predict NeuCo Semantic vs. Low-Level tasks. Diamonds indicate the best-performing block for each model and task. Semantic tasks (left) generally improve and then plateau in intermediate-to-late blocks, whereas low-level tasks (right) show more heterogeneous depth profiles across models. Features are obtained by mean-pooling patch tokens (excluding the CLS token).}
    \label{fig:neuco_layerwise_tasknormalized_grouped_all_models_combined}
\end{figure*}

\begin{figure}[h]
    \centering
    \includegraphics[width=0.45\textwidth]{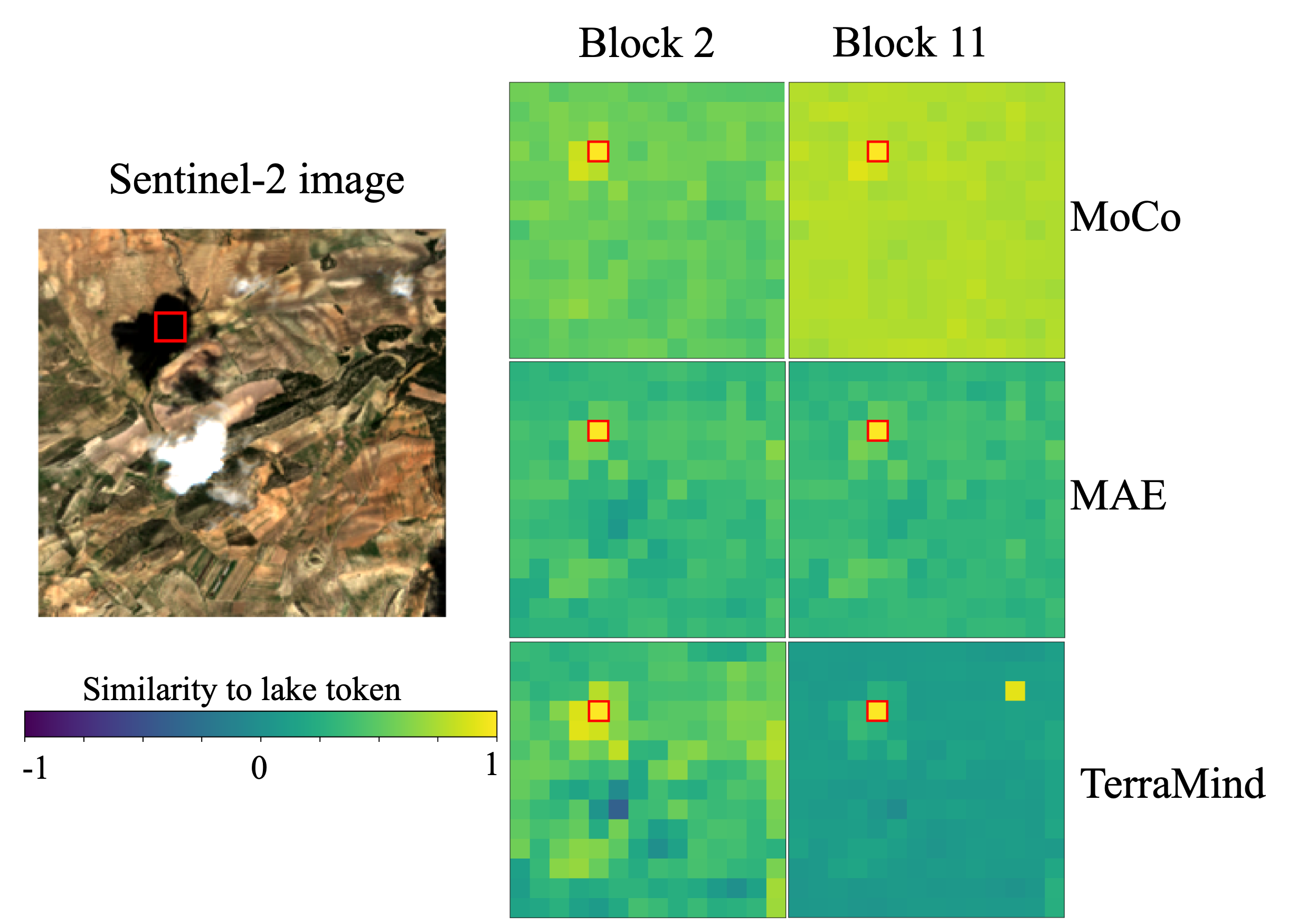}
    \caption{Cosine similarity between a selected lake patch token and all other patch tokens, shown at an early and late block of ViT-B/S. Blocks are 0-indexed.}
    \label{fig:overview}
\end{figure}

\begin{figure}[h]
    \centering
    \includegraphics[width=0.4\textwidth]{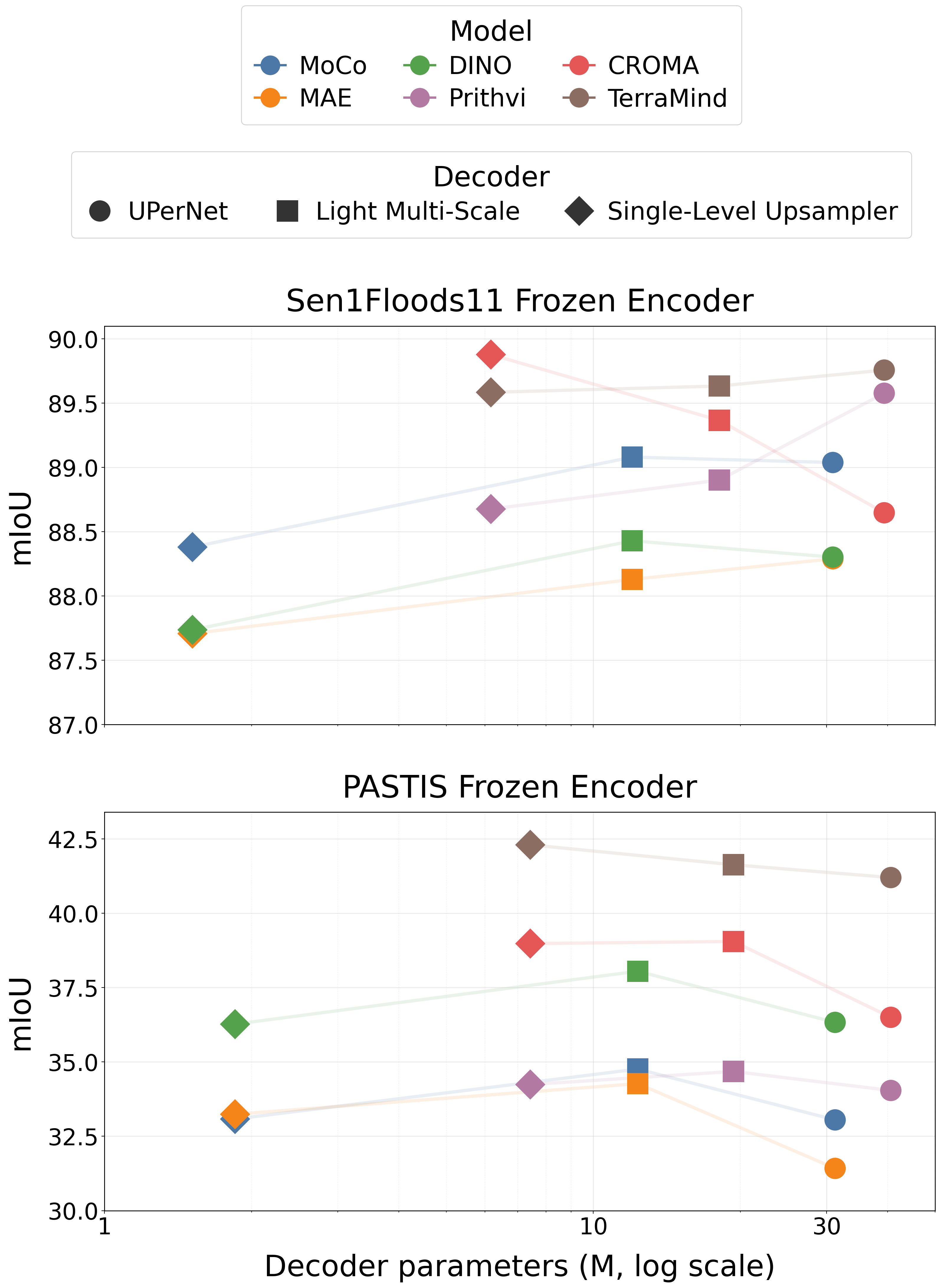}
    \caption{Comparison of Decoder Designs with Fixed Frozen Encoder + L-TAE on PASTIS. For Single-Level Upsampler, the best layer is shown. Within decoder parameter differences are due to ViT-B vs ViT-S backbone differences. }
    \label{fig:scaling-decoder}
\end{figure}

\subsection{Emergence of Representations in GeoFM Vision Encoders}
\subsubsection{Layerwise Linear Probing of Pretrained Representations} \label{layerwise-linear-probe}
In this analysis, we examine the internal workings of GeoFMs through layerwise linear probes to identify how \textbf{low-level} and \textbf{high-level} information is transformed throughout network depth (as defined in Section~\ref{low-vs-high}). Figure~\ref{fig:neuco_layerwise_tasknormalized_grouped_all_models_combined} shows layerwise linear probe performance across the model suite. On semantic tasks, most models improve rapidly over the first few blocks and then plateau in intermediate-to-late layers. 

By contrast, low-level information is distributed more variably over network depth, and we see diverging profiles across the model suite. For several models, the strongest low-level features appear in early blocks.  MAE and Prithvi, reconstruction models, tend to retain useful low-level information deeper than MoCo/DINO, while TerraMind continues improving across depth on most tasks. TerraMind is a notable exception, showing consistent improvement through the full depth of the network for all tasks except for cloud prediction. Interestingly, across all tasks MoCo reaches peak performance at the shallowest blocks across the model suite.

Overall, in this model suite, reconstruction-based models exhibit more gradual improvements across both low-level and semantic tasks, while joint embedding models (MoCo and DINO) show a sharper transition from early low-level phase to later semantic phase. TerraMind uniquely exhibits progressive depthwise representation learning for both low-level and high-level information such that the final representation is best on all tasks except for clouds.  For other models, the most transferable representation for a given task is often found in an intermediate block rather than the final output layer. These insights can be used to improve the downstream transferability of encoders, such as through designing parameter-efficient fine-tuning or domain adaptation modules. Embedding-centric usage can benefit by adaptively selecting layers or aggregating intermediate layer features accordingly.


We present a qualitative illustration of these depthwise differences in Figure~\ref{fig:overview} by comparing patch-token similarity maps for MoCo, MAE, and TerraMind at an early block and the final block. For a selected lake patch, all three models show locally structured similarity in the early layer, but their late-layer behavior differs. TerraMind learns a stronger distinction between lake and non-lake patches, while MAE's similarity map does not change much between the early and late blocks. MoCo, in contrast, shows high similarity across all patches in the image, suggesting that local variations are smoothed out and forgotten deeper in the network.

\subsubsection{How do representations transfer on segmentation tasks?} \label{fine-tune-results}

We next examine how pretrained representations transfer during downstream adaptation on case studies. \textbf{Table~\ref{tab:segmentation_contrast_summary}}  summarizes the average effect of each downstream condition across the six GeoFMs. \textbf{Table~\ref{tab:segmentation_per_model}} reports per-model performance for the Light Multi-Scale (LMS) Decoder, since this design performs better on average ($\boldsymbol{\Delta}$ UPerNet--LMS in \textbf{Table~\ref{tab:segmentation_contrast_summary}}). Results for all encoder, decoder, fine-tuning, and label-availability combinations are in Supp. \textbf{Tables~\ref{tab:sen1floods_decoder_family_matched_full} and ~\ref{tab:segmentation_factorial_clean}}.

\begin{table*}[h]
\centering
\footnotesize
\setlength{\tabcolsep}{4pt}
\renewcommand{\arraystretch}{1.15}
\begin{tabular}{l cc cc cc}
\toprule
& \multicolumn{2}{c}{\multirow{2}{*}{\textbf{PASTIS (L-TAE, 10\% labels)}}} & \multicolumn{4}{c}{\textbf{Sen1Floods11}} \\
\cmidrule(lr){4-7}
& \multicolumn{2}{c}{} & \multicolumn{2}{c}{10\% labels} & \multicolumn{2}{c}{100\% labels} \\
\cmidrule(lr){2-3} \cmidrule(lr){4-5} \cmidrule(lr){6-7}
\textbf{Model} & \textbf{Frozen} & \textbf{Fine-tuned} & \textbf{Frozen} & \textbf{Fine-tuned} & \textbf{Frozen} & \textbf{Fine-tuned} \\
\midrule
MoCo       & 36.18          & 41.39          & \textbf{84.80} & 82.21          & 89.08          & \underline{88.86} \\
MAE        & 35.09          & 40.43          & \underline{84.61} & 81.70       & 88.13          & 88.39 \\
DINO       & \underline{38.70} & 42.90       & 80.32          & \textbf{84.73} & 88.43          & 88.77 \\
Prithvi    & 34.88          & 36.40          & 83.95          & 83.24          & 88.90 & 88.32 \\
CROMA      & 38.61          & \underline{43.31} & 84.41       & 84.27 & \underline{89.37} & 88.80 \\
TerraMind  & \textbf{40.60} & \textbf{45.61} & 83.29          & 83.18          & \textbf{89.63} & \textbf{89.49} \\
\midrule
\textbf{GeoFM Mean $\pm \sigma$} & 37.34 $\pm$ 2.30 & 41.67 $\pm$ 3.13 & 83.56 $\pm$ 1.68 & 83.22 $\pm$ 1.16 & 88.92 $\pm$ 0.57 & 88.77 $\pm$ 0.42 \\
ImageNet ViT-B/16 (13 ch) & 23.87 & 40.95 & 80.66 & \underline{84.36} & 86.12 & 88.84 \\
\bottomrule
\end{tabular}
\caption{GeoFM segmentation mIoU (\%) using the Light Multi-Scale (LMS) decoder. PASTIS results use L-TAE temporal aggregation with 10\% labels. Full per-model results across all decoder and fine-tuning configurations are in Supp. Table~\ref{tab:sen1floods_decoder_family_matched_full}, ~\ref{tab:segmentation_factorial_clean}.}
\label{tab:segmentation_per_model}
\end{table*}

\begin{table}[h]
\centering
\small
\setlength{\tabcolsep}{3pt}
\renewcommand{\arraystretch}{1.15}
\begin{tabular}{llcc}
\toprule
Dataset & Factor & Comparison & $\Delta$ \\
\midrule
PASTIS 
& Temporal 
& Single time-step $\rightarrow$ L--TAE 
& $\mathbf{+23.81}$ \\

PASTIS L--TAE
& Decoder 
& LMS $\rightarrow$ UPerNet 
& $\mathbf{-1.38}$ \\

PASTIS L--TAE
& Training 
& Frozen $\rightarrow$ Fine-tuned 
& $\mathbf{+5.06}$ \\

\midrule
Sen1Floods11
& Labels 
& 10\% $\rightarrow$ 100\% 
& $\mathbf{+5.88}$ \\

Sen1Floods11
& Training 
& Frozen $\rightarrow$ Fine-tuned 
& $\mathbf{-0.19}$ \\

Sen1Floods11
& Decoder 
& LMS $\rightarrow$ UPerNet 
& $\mathbf{-0.33}$ \\
\bottomrule
\end{tabular}
\caption{Impact of downstream pipeline settings on segmentation mIoU. Reported deltas average over the applicable remaining settings and six GeoFMs. LMS denotes Light Multi-Scale. Full mean $\pm$ std values are reported in the supplement.}
\label{tab:segmentation_contrast_summary}
\end{table}



 
\noindent\textbf{At higher label availability, encoder choice is less significant.}

\textbf{Table~\ref{tab:segmentation_per_model}} shows per-model Sen1Floods11 performance at 10\% and 100\% labels using the LMS decoder. At 10\% labels, GeoFM performances are more variable ($\sigma$ = 1.68 frozen, 1.16 fine-tuned), while at 100\% labels the GeoFMs achieve more similar performances ($\sigma$ = 0.57 frozen, 0.42 fine-tuned). The same pattern holds with the UPerNet decoder (Supp. \textbf{Table~\ref{tab:sen1floods_decoder_family_matched_full}}). 

Surprisingly, these case studies show that a supervised ImageNet ViT-B/16 baseline with 13 input channels initialized with cyclical RGB weights is competitive despite no remote sensing pretraining. At 10\% labels, there  is a small difference between encoders, but a fine-tuned ImageNet baseline still matches or exceeds several GeoFMs. GeoFM model selection impacts performance to a larger extent in limited-label settings, regardless of adaptation pipeline. However, ranking is not consistent across models.

\noindent\textbf{Fine-tuning benefits are dataset-dependent.}
On PASTIS with L-TAE, fine-tuning is beneficial for every model (mean +5.06 mIoU gain across both LMS and UPerNet decoders, \textbf{Table ~\ref{tab:segmentation_contrast_summary}}), ranging from +1.52 to +5.34 gain with LMS decoder. On Sen1Floods11, fine-tuning has a much smaller effect, with a mean decrease of -0.19 mIoU averaged across label availability and decoders (\textbf{Table~\ref{tab:segmentation_contrast_summary}}). PASTIS is a more complex task requiring high-level semantic and temporal information to distinguish between 18 crop classes, whereas Sen1Floods11 overall is a simpler task with binary prediction on water vs. no-water. Fine-tuning may be more important on complex or semantic tasks.

We also note that the ImageNet ViT-B Baseline benefits from fine-tuning in every experiment. Fine-tuning on 10\% or 100\% labels is enough for the baseline to become competitive or outperform several GeoFMs.

\begin{figure*}[h]
    \centering

    \begin{subfigure}[t]{0.75\textwidth}
        \centering
        \includegraphics[width=\textwidth,
        trim={2pt 2pt 3pt 2pt},
    clip]{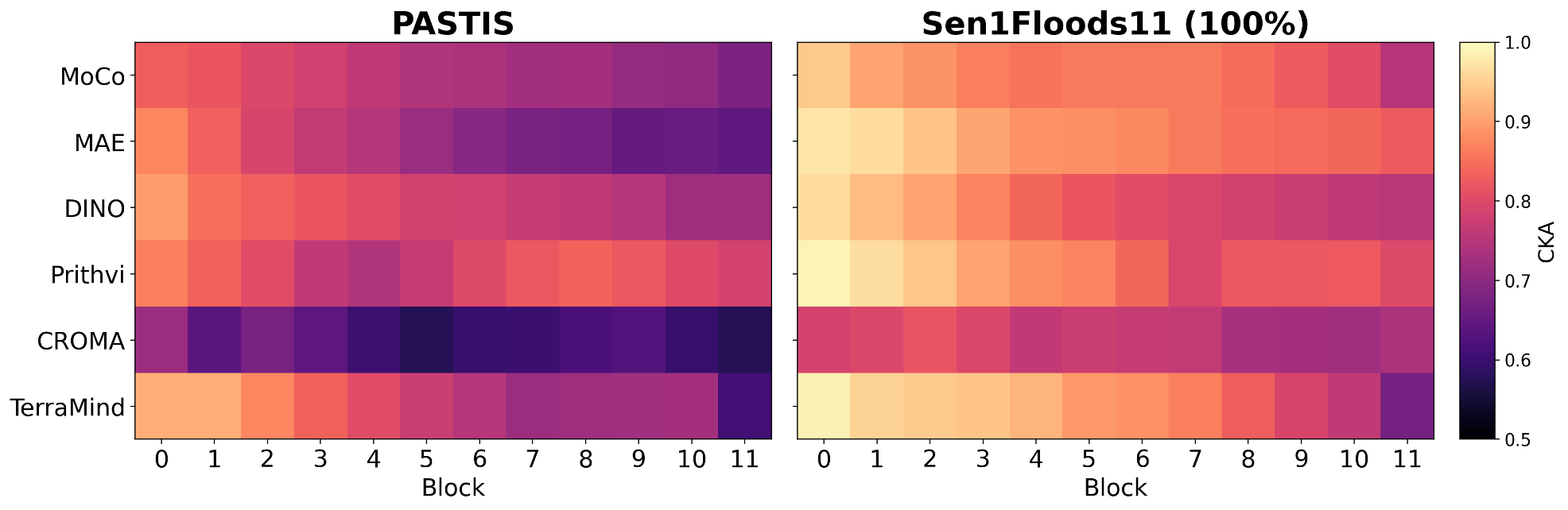}
        \caption{Within-model blockwise CKA similarity between pretrained and fine-tuned encoder representations on PASTIS and Sen1Floods11 datasets. Dark colors indicate areas with more change induced by fine-tuning.}
        \label{fig:cka-a}
    \end{subfigure}


    \begin{subfigure}[t]{0.9\textwidth}
        \centering
        \includegraphics[width=\textwidth,trim={1pt 1pt 3pt 1pt},
    clip]{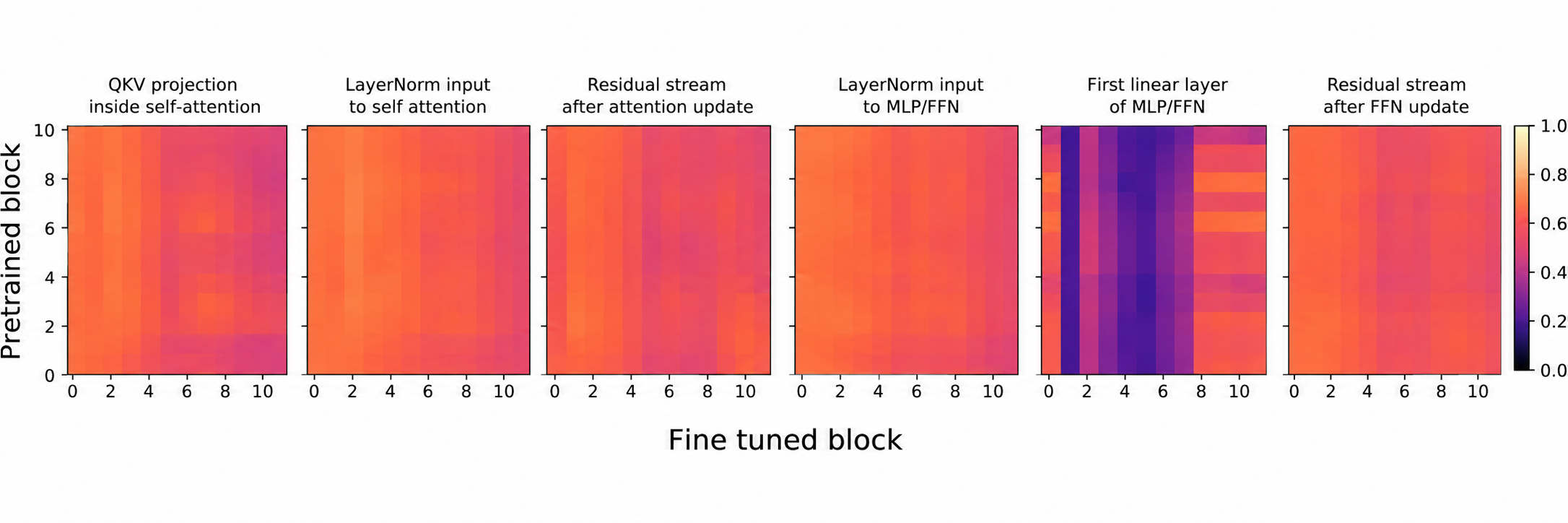}
        \caption{CROMA Pre-trained vs PASTIS Fine-tuned CKA within each layer type. Vertical dark striping indicates that fine-tuning introduces new structure into the representation space that is not present at any depth in the pre-trained encoder. }
        \label{fig:cka-b}
    \end{subfigure}

    \caption{}
    
    \label{fig:block-level shift}
\end{figure*}

\begin{table*}[t]
\centering
\small
\setlength{\tabcolsep}{4pt}
\begin{tabular}{lccccc}
\toprule
Setting 
& UPerNet 
& Block 3 
& Block 5 
& Block 7 
& Block 11 \\
\midrule
PASTIS 10\% 
& $35.24 \pm 3.08$ 
& $35.59 \pm 2.85$ 
& $35.90 \pm 3.19$ 
& \textbf{36.00 $\pm$ 3.62}
& $35.42 \pm 3.00$ \\

Sen1Floods11 10\% 
& $82.78 \pm 2.65$ 
& $83.14 \pm 3.16$ 
& \textbf{84.43 $\pm$ 1.93}
& $83.09 \pm 2.12$ 
& $83.87 \pm 1.09$ \\

Sen1Floods11 100\% 
& \textbf{88.94 $\pm$ 0.63} 
& $88.35 \pm 0.77$ 
& $88.54 \pm 0.87$ 
& $88.60 \pm 0.92$ 
& $88.51 \pm 0.80$ \\
\bottomrule
\end{tabular}
\caption{Frozen-encoder layerwise segmentation results using the single-level upsampler compared with the multi-scale UPerNet baseline. Values report mean mIoU $\pm$ standard deviation across the six GeoFMs. Full per-model layerwise results are reported in the supplement.}
\label{tab:layerwise_seg_summary}
\end{table*}

\noindent\textbf{Lighter single-level decoders are competitive with multi-scale fusion.}

Figure~\ref{fig:scaling-decoder} compares all three decoder designs on PASTIS and Sen1Floods11 with frozen encoders. Our single-level upsampler, which uses only one encoder feature-level and applies progressive transposed convolutions, often matches or outperforms both the lightweight multi-scale fusion and the heavyweight UPerNet despite using 5 or 20$\times$ fewer decoder parameters than UPerNet (Table~\ref{tab:decoder_summary}). The layerwise segmentation results in \textbf{Table~\ref{tab:layerwise_seg_summary}} show that the single-level upsampler is often best using intermediate layers (Block 5 and 7) rather than the final output features. On Sen1Floods11 with 100\% of the training data, UPerNet tends to outperform the single-level decoder. Per-model results are shown in Supp. ~\ref{tab:block_results}. These results suggest that the standard UPerNet multi-scale fusion decoder may not align with how GeoFMs organize information over depth.


\noindent\textbf{Fine-Tuning revises layers differently across models and shifts are concentrated in distinct sub-layers.}
Figure ~\ref{fig:cka-a} shows CKA similarity between matched pretrained and fine-tuned sublayers, averaged within each ViT block. Full CKA matrices are shown in Supp. Figure ~\ref{fig:full-cka}. GeoFMs exhibit distinct patterns in layer representation shifts under fine-tuning regarding block representations over depth and sub-layers within blocks.

\textit{Representation shift is distributed over depth.} Most models' shift  increases over depth. The exceptions are Prithvi, which has more shift in intermediate layers than early and late layers, and CROMA, which undergoes substantial shift throughout the network on both tasks.  TerraMind representation shifts are concentrated in the final ViT Block, signaling more generalizable early representations.

On PASTIS, all models exhibit performance gains during fine-tuning (\textbf{Table ~\ref{tab:segmentation_per_model}}). On Sen1Floods11, models undergo less shift overall than on PASTIS during fine-tuning, and the shift is more condensed to later layers. DINO is the only model that significantly benefits from fine-tuning on Sen1Floods11.

\textit{Representation shift is concentrated in an MLP layer.} CROMA exhibits representation shift concentrated within the first linear layer of the MLP modules (Figure ~\ref{fig:cka-b}).  We found that all other encoders also exhibit maximum representation change in this layer-type (Supp. Fig~\ref{fig:cka-supp}b), although their shifts are not as large as CROMA.  This analysis indicates that certain sub-layers may be more suitable for downstream adaptation. 

Related work in computer vision empirically demonstrates that selectively fine-tuning chosen modules (attention, MLP, and LayerNorm) or depths (early vs late ViT blocks) can match or outperform full fine-tuning, depending on the dataset and task \cite{ye2023partialfinetuningsuccessorfinetuning, lee2023surgicalfinetuningimprovesadaptation, tan2025exploitinglayernormalizationfinetuning}. 



\noindent\textbf{Pretraining Scale versus Transfer Reliability}
Tables~\ref{tab:model_suite_objectives}, ~\ref{tab:model_suite_scale}: Our model suite spans nearly 50$\times$ parameter counts in the pretraining framework (MoCo at 42M params to TerraMind at ~2B), and the three general-purpose models use ViT-S (23M) , whereas the geospatial models use ViT-B (86M) backbones. The pretraining datasets also span 1M samples to 9M samples with 7+ modalities, and limited geographic coverage to global coverage, with TerraMind trained on the largest dataset with global coverage. TerraMind is strong across each evaluation, achieving first or second rank on most tasks; showing it is rather robust to downstream task transfer. However, we see small general-purpose models and the ImageNet supervised baseline remain competitive on Sen1Floods11 dataset. TerraMind appears to be a first strong indication that scaling up GeoFM pretraining improves robustness and generalization, but there are still task-specific challenges not addressed by scaling.  


\noindent\textbf{Training Efficiency.}
Changing the decoder design has modest impacts on training time (Supp. Table~\ref{tab:pastis_runtime_summary}). From UPerNet to Single-Level Upsampler, parameters are reduced by 82-93\% and training time is reduced by 26\%. The major compute savings come from using the Frozen setup compared to Fine-tuning, which results in a 52\% reduction in training time on average across decoders. Overall, the lightweight decoders reduce parameters substantially and we see the lightest decoder often outperforms the heaviest decoder.

\subsection{ImageNet Baseline} \label{imagenet}
Prior works benchmarking GeoFMs use varying baselines. \cite{marsocci2025pangaeaglobalinclusivebenchmark} evaluates GeoFMs against fully-supervised baselines, finding that GeoFMs struggle to beat UNet and ViT supervised baselines on dense prediction. Works using ImageNet baselines either use RGB-only \cite{simumba2026geobench2performancecapabilityrethinking}, use multi-spectral channels with cyclical RGB weights like ours but only evaluate frozen-encoder setups on classification \cite{corley2024revisiting, corley2025landsat} or only evaluate fine-tuning on classification tasks \cite{lacoste2023geo} using RGB channel initialization and random for multi-spectral channels. Our work contributes an analysis of multi-spectral ImageNet baseline performance on classification and regression tasks in the frozen setup along with frozen vs fine-tuned settings on segmentation tasks. These results show that the baseline is not strongly competitive with modern GeoFMs on the frozen setups across task types, but fine-tuning on PASTIS 10\% labels and Sen1Floods11 with 10\% or 100\% labels makes the baseline competitive. On Sen1Floods11 with 10\% labels, the ImageNet baseline ranks 2nd place. 

\section{Discussion and Conclusion}
\label{sec:conclusion}

Across downstream prediction tasks, we find that model rankings are inconsistent, with the exception of TerraMind usually ranking close to top. Across classification, regression, and segmentation tasks, GeoFM rankings are highly sensitive to both task type and downstream transfer setup. Our results show that SSL models organize task-relevant information differently across depth. Semantic tasks generally become linearly accessible in intermediate-to-late layers, whereas low-level sensor-related tasks exhibit more heterogeneous depth profiles. 

Our segmentation case studies further show that downstream adaptation method can be more impactful on performance than the specific foundation model encoder used. Additionally, common multi-scale fusion decoders do not consistently outperform lighter single-level decoders, suggesting that standard dense-prediction heads may not be well matched to how GeoFMs organize information over depth or to the requirements of dense geospatial task settings. With fine-tuning on segmentation tasks, the supervised ImageNet baseline (adapted to 13-channels) and general SSL models trained on remote sensing data are competitive with geospatial-specific models trained with substantially larger pretraining frameworks and datasets.

Finally, CKA analysis shows how fine-tuning rewrites GeoFMs. Representation shifts differ across models and datasets, and the strongest changes can localize to specific sub-modules rather than being spread evenly throughout the network. This can guide downstream transfer methods for GeoFMs.



This work is a step towards interpreting representation learning in GeoFMs and linking it with downstream adaptation. Regarding GeoFM development, we note that on our examined tasks, the largest GeoFM frameworks do not always outperform much smaller general SSL frameworks or the ImageNet baseline. These results are observed across classification and regression tasks (using single layer probes or kNN and frozen backbones), as well as on segmentation tasks using light to heavyweight decoders and frozen vs fine-tuning. These findings highlight that current downstream transfer pipelines may not align well with GeoFM representations. Our work highlights the need for carefully designed prediction heads which can enhance downstream adaptation.

Given these findings, we encourage future work into domain-specific decoders that can adapt to how information is structured over encoder depth, such as adaptive layer selection, embedding aggregation, or multimodal fusion. Furthermore, patch size of ViT limits the spatial granularity of tokens and we see consistent inaccuracies in segmenting narrow or small pixel areas (Supp Fig \ref{fig:floods}). Decreasing patch size scales up compute requirements drastically, thus development of task-specific adaptation modules that encourage retaining within-patch structure may be valuable for increasing dense prediction accuracy.

 As more global labeled and multi-source datasets become available, we encourage further work into GeoFM interpretation to uncover domain-specific behaviors related to representation learning, such as the impacts of multimodal pretraining, transfer performance to new geographic regions, image resolutions, or robustness to geospatial artifacts and environmental noises. Interpreting these internal mechanisms can provide guidance for GeoFM development and the design of more efficient adaptation strategies.  

{
    \small
    \bibliographystyle{ACM-Reference-Format.bst}
    \bibliography{sample-base.bib}
}
\clearpage
\appendix
\onecolumn
\clearpage

\section{Supplementary}
\label{sec:supplementary}

\begin{table}[h]
\centering
\scriptsize
\setlength{\tabcolsep}{6pt}
\begin{tabular}{llc|cccc}
\toprule
Setting & Encoder (Frozen) & Multi-Scale ($\Delta$ Best Block) & Block 3 & Block 5 & Block 7 & Block 11 \\
\midrule
\multirow{6}{*}{PASTIS 10\%}
& MoCo      & 33.05 \textcolor{red}{(-0.03)}            & 32.71 & \textbf{33.08} & 32.23 & 31.61 \\
& MAE       & 31.42 \textcolor{red}{(-1.82)}            & 32.66 & 32.61 & 33.10 & \textbf{33.24} \\
& DINO      & 36.33 \textcolor{green!60!black}{(+0.06)} & 35.92 & \textbf{36.27} & 35.96 & 36.23 \\
& Prithvi   & 34.04 \textcolor{red}{(-0.20)}            & 33.87 & 34.16 & \textbf{34.24} & 33.93 \\
& CROMA     & 36.50 \textcolor{red}{(-2.48)}            & 38.48 & \textbf{38.98} & 38.96 & 38.23 \\
& TerraMind & 40.11 \textcolor{red}{(-1.41)}            & 39.90 & 40.32 & \textbf{41.52} & 39.30 \\
\midrule
\multirow{6}{*}{Sen1Floods11 10\%}
& MoCo      & 84.36 \textcolor{red}{(-1.00)}            & \textbf{85.36} & 85.09 & 85.03 & 84.72 \\
& MAE       & 78.63 \textcolor{red}{(-5.75)}            & 78.34 & 82.31 & 82.94 & \textbf{84.38} \\
& DINO      & 80.87 \textcolor{red}{(-2.20)}            & 82.64 & \textbf{83.07} & 82.22 & 82.66 \\
& Prithvi   & 83.65 \textcolor{green!60!black}{(+0.37)} & 82.81 & \textbf{83.28} & 81.54 & 82.85 \\
& CROMA     & 86.07 \textcolor{red}{(-1.50)}            & 87.18 & \textbf{87.57} & 85.29 & 84.45 \\
& TerraMind & 83.07 \textcolor{red}{(-2.17)}            & 82.49 & \textbf{85.24} & 81.47 & 84.16 \\
\end{tabular}
\caption{Frozen encoder layerwise segmentation results using single-level upsampler compared with multi-scale UPerNet. }
\label{tab:block_results}
\end{table}

\begin{table}[h]
\centering
\scriptsize
\setlength{\tabcolsep}{4.8pt}
\renewcommand{\arraystretch}{1.14}
\begin{tabular}{lcccc@{\hspace{7mm}}cccc}
\toprule
& \multicolumn{4}{c}{\textbf{Light Multi-Scale (12-18M params)}} 
& \multicolumn{4}{c}{\textbf{UPerNet (31--39M params)}} \\
\cmidrule(lr){2-5} \cmidrule(lr){6-9}
& \multicolumn{2}{c}{\textbf{10\% labels}} 
& \multicolumn{2}{c}{\textbf{100\% labels}}
& \multicolumn{2}{c}{\textbf{10\% labels}} 
& \multicolumn{2}{c}{\textbf{100\% labels}} \\
\cmidrule(lr){2-3} \cmidrule(lr){4-5}
\cmidrule(lr){6-7} \cmidrule(lr){8-9}
\textbf{Model}
& \textbf{Frozen} & \textbf{Fine-tuned}
& \textbf{Frozen} & \textbf{Fine-tuned}
& \textbf{Frozen} & \textbf{Fine-tuned}
& \textbf{Frozen} & \textbf{Fine-tuned} \\
\midrule
MoCo
& \textbf{84.80} & 82.21 & 89.08 & \underline{88.86}
& \underline{84.36} & \textbf{87.12} & 89.04 & 88.94 \\
MAE
& \underline{84.61} & 81.70 & 88.13 & 88.39
& 78.63 & 82.11 & 88.29 & 88.28 \\
DINO
& 80.32 & \textbf{84.73} & 88.43 & 88.77
& 80.87 & 84.28 & 88.30 & \underline{89.45} \\
Prithvi
& 83.95 & 83.24 & 88.90 & 88.32
& 83.65 & 74.74 & \underline{89.58} & 89.32 \\
CROMA
& 84.41 & 84.27 & \underline{89.37} & 88.80
& \textbf{86.07} & 84.30 & 88.65 & 87.84 \\
TerraMind
& 83.29 & 83.18 & \textbf{89.63} & \textbf{89.49}
& 83.07 & 82.44 & \textbf{89.76} & \textbf{89.80} \\
\midrule
\textbf{GeoFM Mean$\pm$ Std}
& $83.56 \pm 1.68$ & $83.22\pm 1.16$ 
& $88.92\pm 0.57$ & $88.77\pm 0.42$
& $82.78\pm 2.65$ & $82.50\pm 4.20$ 
& $88.94\pm 0.63$ & $88.94\pm 0.75$ \\
\midrule
\textbf{ImageNet ViT-B/16 (13 ch)}
& 80.66 & \underline{84.36} & 86.12 & 88.84
& 82.89 & \underline{84.60} & 86.28 & 87.96 \\
\bottomrule
\end{tabular}
\caption{Segmentation mIoU (\%) on Sen1Floods11 for both Light Multi-Scale and UPerNet. Parameter counts show only decoder parameter counts, although fine-tuned settings include encoder parameters.} 
\label{tab:sen1floods_decoder_family_matched_full}
\end{table}

\begin{table}[h]
\centering
\scriptsize
\setlength{\tabcolsep}{5.5pt}
\renewcommand{\arraystretch}{1.15}

\begin{tabular}{
l
S S S S
@{\hspace{7mm}}
S S S S
}
\toprule
& \multicolumn{4}{c}{\textbf{PASTIS Single-Timestep}} & \multicolumn{4}{c}{\textbf{PASTIS Temporal Aggregation (L--TAE)}} \\
\cmidrule(lr){2-5} \cmidrule(lr){6-9}
\textbf{Model}
& \multicolumn{2}{c}{\textbf{Light Multi-Scale (12–18M params)}}
& \multicolumn{2}{c}{\textbf{UPerNet (31–39M params)}}
& \multicolumn{2}{c}{\textbf{Light Multi-Scale  (12–18M params)}}
& \multicolumn{2}{c}{\textbf{UPerNet (31–39M params)}} \\
\cmidrule(lr){2-3} \cmidrule(lr){4-5}
\cmidrule(lr){6-7} \cmidrule(lr){8-9}
& {\textbf{Frozen}} & {\textbf{Fine-tuned}}
& {\textbf{Frozen}} & {\textbf{Fine-tuned}}
& {\textbf{Frozen}} & {\textbf{Fine-tuned}}
& {\textbf{Frozen}} & {\textbf{Fine-tuned}} \\
\midrule
MoCo
  & 12.67 & 15.09 & 14.29 & 14.67 & 36.18 & 41.39 & 33.05 & 40.91 \\
MAE
  & 13.53 & 15.46 & 13.17 & 14.57 & 35.09 & 40.43 & 31.42 & 40.53 \\
DINO
  & 14.40 & \underline{16.62} & \underline{14.56} & \underline{15.56}
  & \underline{38.70} & 42.90 & 36.33 & 41.82 \\
Prithvi
  & 10.29 & 13.20 & 13.11 & 13.72 & 34.88 & 36.40 & 34.04 & 34.89 \\
CROMA
  & \underline{14.66} & 15.15 & 14.54 & 15.32
  & 38.61 & \underline{43.31} & \underline{36.50} & \underline{42.58} \\
TerraMind
  & \textbf{19.67} & \textbf{19.90} & \textbf{17.50} & \textbf{18.56}
  & \textbf{40.60} & \textbf{45.61} & \textbf{40.11} & \textbf{45.38} \\
\midrule
\textbf{GeoFM Mean$\pm$ Std}
  & {$14.20 \pm 3.11$} & {$15.90 \pm 2.25$}
  & {$14.53 \pm 1.60$} & {$15.40 \pm 1.68$}
  & {$37.34 \pm 2.30$} & {$41.67 \pm 3.13$}
  & {$35.24 \pm 3.08$} & {$41.02 \pm 3.46$} \\
\midrule
\textbf{ImageNet ViT-B/16 (13 ch)}
& \multicolumn{1}{c}{---}
& \multicolumn{1}{c}{---}
& \multicolumn{1}{c}{---}
& \multicolumn{1}{c}{---}
& 23.87 & 40.95 & 22.64 & 41.11 \\
\addlinespace[1pt]
\bottomrule
\end{tabular}
\caption{Segmentation mIoU (\%) on PASTIS optical data using 10\% labels. The parameter-heavy UPerNet decoder  yields little or negative gain compared to LMS decoder.}
\label{tab:segmentation_factorial_clean}
\end{table}

\begin{table}[h]
\centering
\scriptsize
\setlength{\tabcolsep}{6pt}
\renewcommand{\arraystretch}{1.12}
\begin{tabular}{lccc ccc}
\toprule
& \multicolumn{3}{c}{\textbf{Frozen}} & \multicolumn{3}{c}{\textbf{Fine-tuned}} \\
\cmidrule(lr){2-4} \cmidrule(lr){5-7}
\textbf{Decoder} &
\textbf{Params (M)} &
\textbf{Total Runtime} &
\textbf{Min--Max} &
\textbf{Params (M)} &
\textbf{Total Runtime} &
\textbf{Min--Max} \\
\midrule
UPerNet (ch=512)
& 31--41
& 5:55
& 0:44--1:32
& 54--128
& 12:02
& 1:05--3:49 \\

Light Multi-Scale (ch=512)
& 12--19
& 5:19
& 0:34--1:26
& 35--107
& 10:55
& 1:06--3:20 \\

Light Multi-Scale (ch=128)
& 3--9
& 5:05
& 0:37--1:25
& 26--97
& 10:46
& 1:03--3:20 \\

Single-Level Upsampler
& 2--7
& 4:20
& 0:29--1:14
& ---
& ---
& --- \\
\bottomrule
\end{tabular}
\caption{Training time for experiments on all six models using different decoders on PASTIS 10\% with an NVIDIA RTX A6000. }
\label{tab:pastis_runtime_summary}
\end{table}

\begin{figure}[h]
    \centering

    \begin{subfigure}{0.70\textwidth}
        \centering
        \includegraphics[width=0.75\textwidth,
        trim={2pt 2pt 3pt 2pt},
    clip]{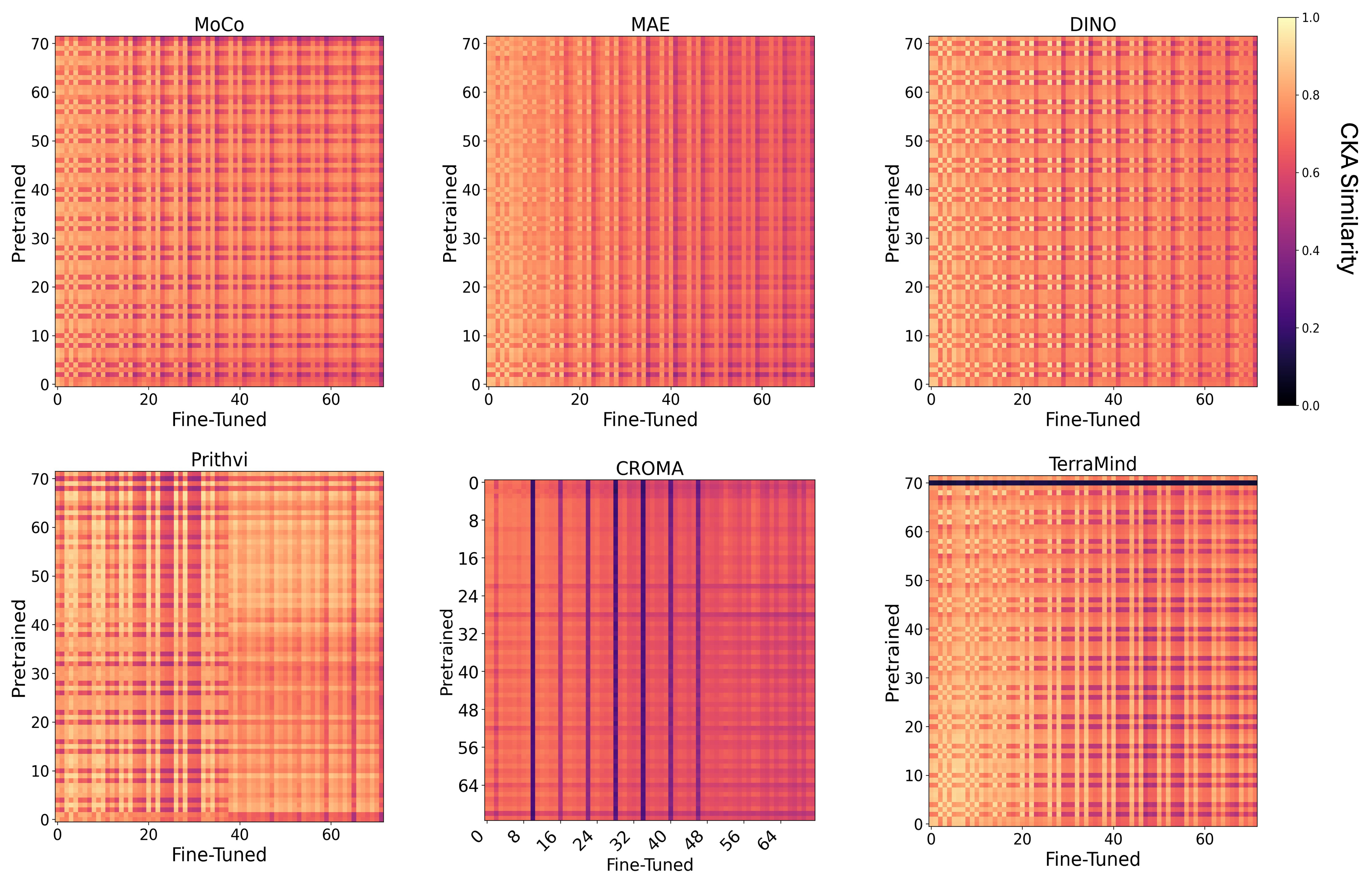}
        \label{fig:full-cka}
    \caption{Within-model layerwise representation similarity between pretrained and fine-tuned encoders on PASTIS 10\% dataset, using L-TAE+UPerNet modules. Vertical dark bands indicate that fine-tuning introducing new structure into representation spaces.}
    \label{fig:full-cka}
    \end{subfigure}

    \vspace{0.75em}

    \begin{subfigure}{0.80\textwidth}
        \centering
    \includegraphics[width=0.8\textwidth,
        trim={3pt 3pt 3pt 3pt},
    clip]{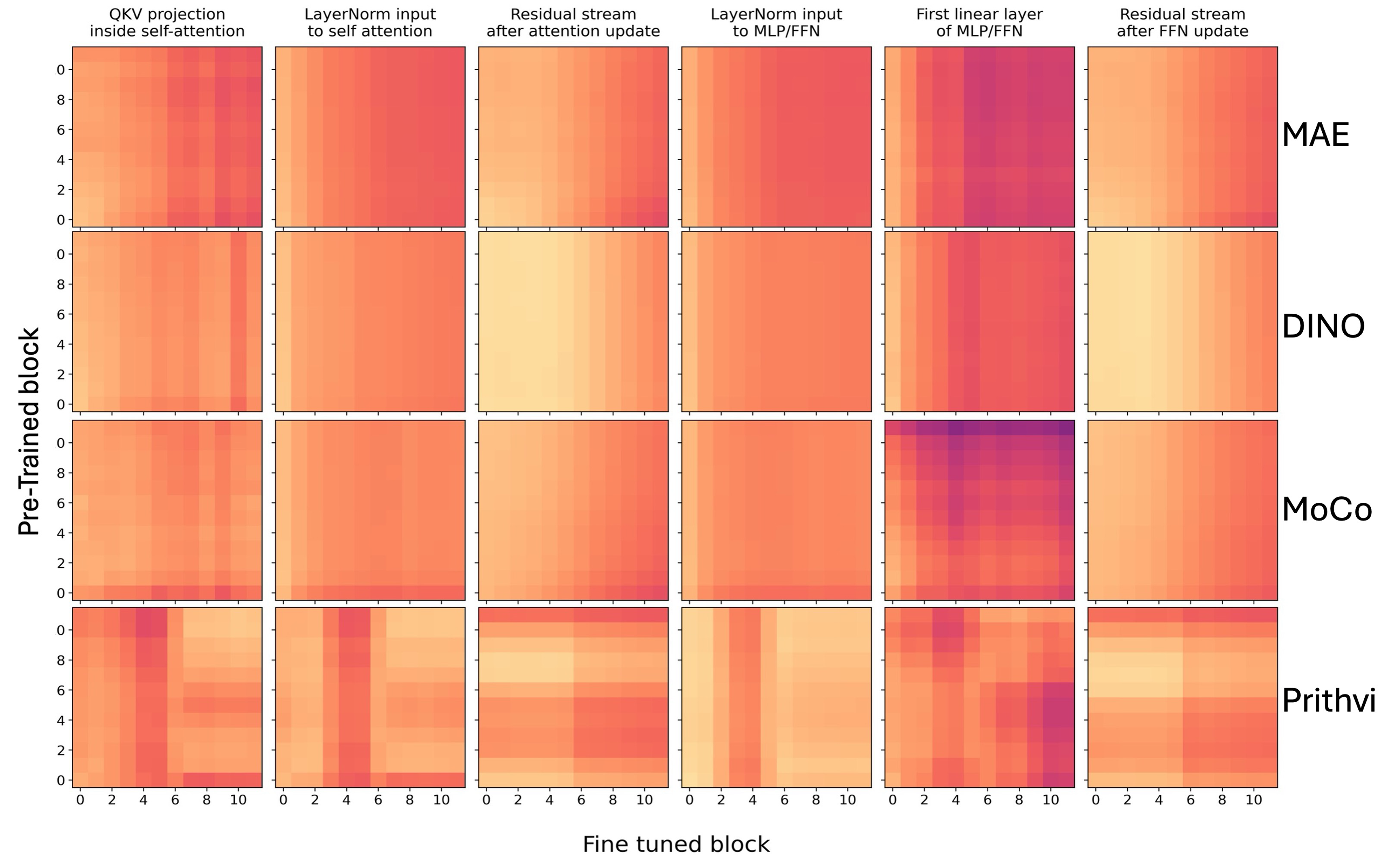}
        \label{fig:supp-cka-b}
    \caption{Per-layer hook CKA between pretrained and PASTIS fine-tuned for selected models. Changes in representation structure are concentrated in the MLP first linear layer across encoders. Color scale is same as part (a).}
    \end{subfigure}

    \caption{
    Representational analysis of fine-tuned segmentation models.
    }
    \label{fig:cka-supp}
\end{figure}

\begin{figure}[t]
    \centering
    \includegraphics[width=0.6\textwidth]{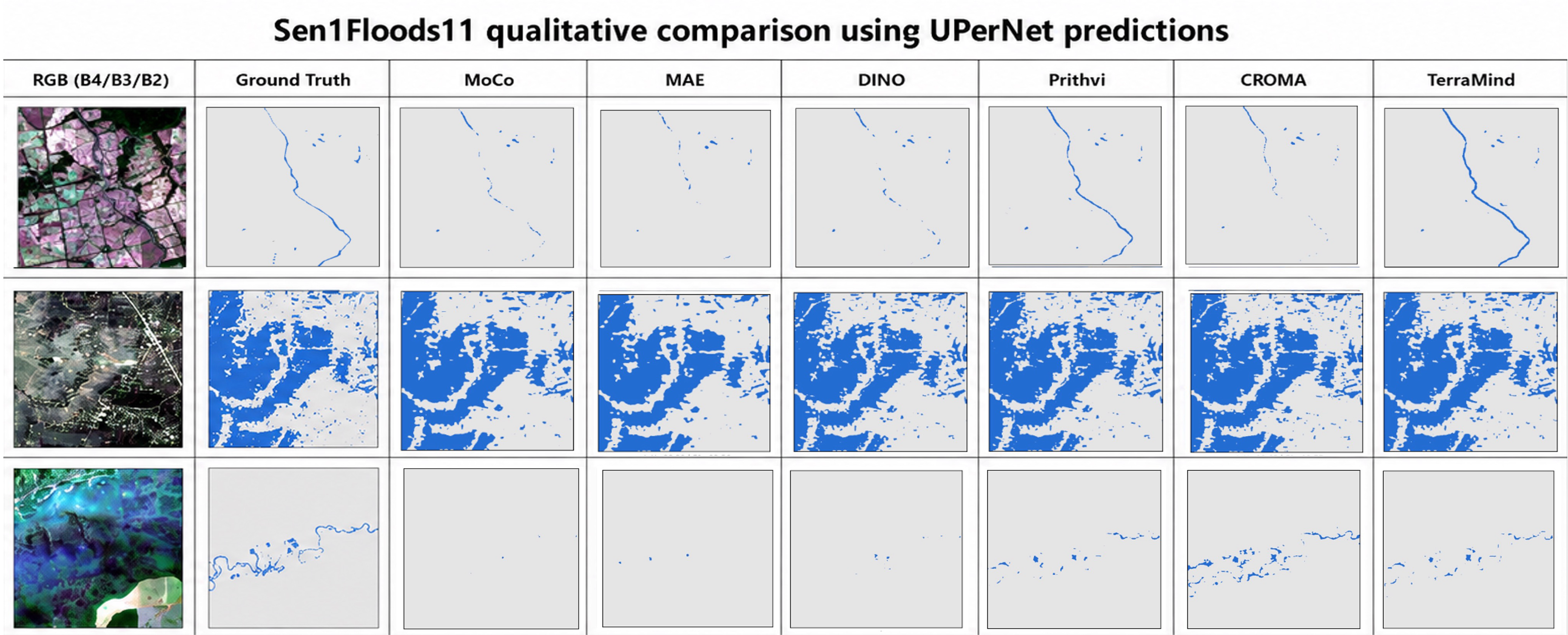}
        \label{fig:floods-qual}
    \caption{Sen1Floods11 100\% binary segmentation masks using fine-tuned encoders + UPerNet.}
    \label{fig:floods}
\end{figure}

\end{document}